\documentclass[11pt,a4paper]{article}
\usepackage{times,latexsym}
\usepackage{url}
\usepackage[T1]{fontenc}

\usepackage[acceptedWithA]{tacl2021v1}

\usepackage{xspace,mfirstuc,tabulary}

\newif\iftaclinstructions
\taclinstructionsfalse 
\iftaclinstructions
\renewcommand{\confidential}{}
\renewcommand{\anonsubtext}{(No author info supplied here, for consistency with
TACL-submission anonymization requirements)}
\newcommand{\instr}
\fi

\iftaclpubformat 

\else

\fi

\usepackage{graphicx}
\usepackage{multirow}
\usepackage{amsmath}
\usepackage{amssymb}
\usepackage{subcaption}
\usepackage{array}
\newcolumntype{?}{!{\vrule width 2pt}}
\mathchardef\mhyphen="2D 
\DeclareMathOperator*{\argmax}{arg\,max}

\usepackage{arydshln}
\usepackage{xcolor}
\definecolor{OliveGreen}{rgb}{0,0.6,0}

\title{T3L: Translate-and-Test Transfer Learning\\for Cross-Lingual Text Classification}

\vspace{12pt}

\author{Inigo Jauregi Unanue$^{\dagger,\star}$,  Gholamreza Haffari$^\bigtriangledown$ \textbf{and} Massimo Piccardi$^\star$ \\
  $^\dagger$RoZetta Technology, Australia \\
  \texttt{inigo.jauregi@rozettatechnology.com} \\
  $^\bigtriangledown$Monash University, Australia \\
  \texttt{gholamreza.haffari@monash.edu}\\
  $^\star$University of Technology Sydney, Australia \\
  \texttt{massimo.piccardi@uts.edu.au} \\}

\begin{document}
\maketitle
\begin{abstract}
Cross-lingual text classification leverages text classifiers trained in a high-resource language to perform text classification in other languages with no or minimal fine-tuning (zero/few-shots cross-lingual transfer). 
Nowadays, cross-lingual text classifiers are typically built on large-scale, multilingual language models (LMs) pretrained on a variety of languages of interest. However, the performance of these models vary significantly across languages and classification tasks, suggesting that the superposition of the language modelling and classification tasks is not always effective. 
For this reason, in this paper we propose revisiting the classic ``translate-and-test'' pipeline to neatly separate the translation and classification stages. The proposed approach couples 1) a neural machine translator translating from the targeted language to a high-resource language, with 2) a text classifier trained in the high-resource language, but the neural machine translator generates ``soft'' translations to permit end-to-end backpropagation during fine-tuning of the pipeline.
Extensive experiments have been carried out over three cross-lingual text classification datasets (XNLI, MLDoc and MultiEURLEX), with the results showing that the proposed approach has significantly improved performance over a competitive baseline.
\end{abstract}


\section{Introduction}
\label{sec:introduction}

Pretrained language models (LMs) have become ubiquitous in natural language processing (NLP) in recent years \cite{peters-etal-2018-deep,radford2018improving,devlin-etal-2019-bert,raffel2020t5,lewis-etal-2020-bart}. These large, transformer-based \cite{vaswani2017attention} deep networks are often pretrained over hundreds of gigabytes or even terabytes of textual data in an unsupervised manner using masked language model training objectives. After pretraining, LMs are typically fine-tuned with supervised datasets for a variety of downstream tasks (e.g., text classification, natural language generation, question answering), regularly achieving state-of-the-art results.

Multilingual LMs \cite{liu2020multilingual,conneau2020unsupervised,pfeiffer2022lifting} -- LMs trained with monolingual corpora from multiple languages (e.g., 25, 50, 100) -- are typically used to extend the downstream tasks to multilingual and cross-lingual scenarios. In their cross-lingual application, multilingual LMs are fine-tuned for a given downstream task using training data in a high-resource language (e.g., English), and then used for inference in another, less-resourced target language, either with no additional fine-tuning (zero-shot) or minimal fine-tuning (few-shot) in the target language.

\begin{figure*}[t!]
	\centering
	\includegraphics[width=0.7\linewidth]{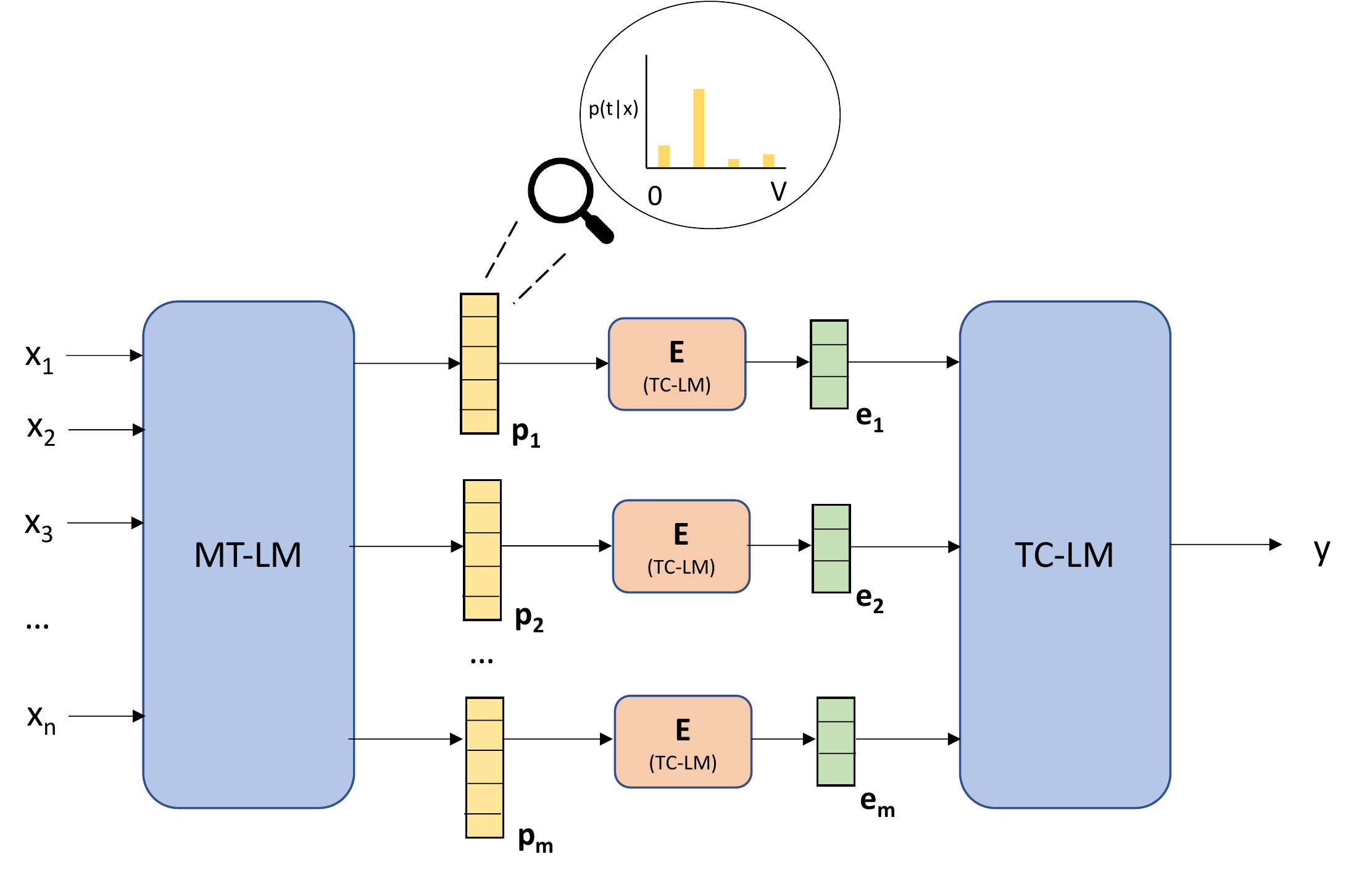}
	\caption{Model architecture for the T3L framework, with the machine translation language model (MT-LM) and the text classification language model (TC-LM) in evidence. $\{ x_1, \dots, x_n\}$ are the tokens of the input sentence in the target language, and  $\mathbf{\{p_1, \dots, p_m\}}$ are the predicted probability distributions over vocabulary $V$ for each token in the translation. $\textbf{E}$ is the TC-LM embedding layer which is used to obtain the expected embeddings, $\mathbf{\{e_1 \ldots e_m\}}$, that are used as input to the TC-LM to predict output label $y$.}
	\label{fig:t3l_model}
\end{figure*}

However, the cross-lingual performance of multilingual LMs tends to vary significantly across languages and downstream tasks, fundamentally because of the deeply uneven training resources and the intrinsic linguistic differences \cite{kreutzer2022quality}. 
While it is possible, in principle, to expand the pretraining with additional monolingual data in the target language,
the computational requirements are often remarkably expensive. On top of that, many low-resource languages do not avail of sufficient amounts of monolingual data in the first place \cite{joshi2020state}. More so, recent research has shown that multilingual LMs can suffer from the \textit{curse of multilinguality} \cite{pfeiffer2022lifting}, which means that at a parity of model capacity, training an LM with too many languages may cause language interference and degrade its cross-lingual performance. In addition, there is room to speculate that the superposition of a downstream task to the multilingual embedding in the same encoder may lead to further interference.

For these reasons, in this paper we propose approaching 
cross-lingual text classification by combining explicit translation from the target language to a high-resource language (i.e., English) with classification in the high-resource language. This approach is known as \textit{translate-and-test} in the literature \cite{conneau2020unsupervised}, and in the recent years it has mainly been regarded as a mere baseline for comparison. However, it offers principled advantages: 1) the possibility to reuse the wealth of existing resources for machine translation, i.e. pretrained MT models, accumulated over decades of research and publicly shared in repositories such as Hugging Face\footnote{\href{https://huggingface.co/}{https://huggingface.co/}}; and 2) exploiting robust text classifiers trained in high-resource languages. On the other hand, a cascaded approach such as translate-and-test is admittedly vulnerable to errors from both the translation stage and the possible semantic mismatch between the LM employed for translation and that used for classification. 
For this reason, we also propose fine-tuning the two modules end-to-end with a few-shot classification training data in the target language. To this aim, the proposed approach -- named T3L for translation-and-test transfer learning - generates \textit{soft} translations \cite{jauregi2021berttune} to ensure the continuity of the network and allow backpropagation end-to-end\footnote{To the best of our knowledge, this is the first proposal of joint fine-tuning of the translation and classification modules in the translate-and-test approach.}. Our experimental results show that the joint fine-tuning has improved the coupling of the two models and significantly contributed to the final classification accuracy. In addition, unlike LM pretraining which is extremely computationally-intensive, fine-tuning this model has been easily manageable and has been carried out with a single GPU for all datasets.
Overall, the main contributions of our paper are:

\begin{itemize}


    \item 
    A translate-and-test pipeline that leverages contemporary language models and resources.
    
    \item 
    An end-to-end pipeline that leverages ``soft'' translations (i.e., predictions of differentiable token embeddings rather than tokens), allowing  fine-tuning the translation and classification modules jointly.
    
    \item A comparative experimental evaluation over three, diverse cross-lingual text classification datasets (XNLI, MLDoc and MultiEURLEX), showing the competitive performance of the proposed approach.
    
    \item An extensive sensitivity, ablation and qualitative analysis, including the impact of the translation quality on the classification accuracy.
    
\end{itemize}

\section{Related Work}
\label{sec:related_work}

Cross-lingual text classification dates back to at least the seminal work of \citet{Bel2003}. Their approach was based on machine translation from the targeted language to English, but was restricted to a selected terminology per class since full-document translation at the time was still very expensive and inaccurate. While much of the early work kept on being based on explicit or latent (e.g., \cite{Wu2008}) machine translation, the advent of word embeddings paved the way for approaches based on multilingual representations of text, which leveraged distributional similarities across languages  \citep{Klementiev2012}. Eventually, the rise of the masked language models \cite{devlin-etal-2019-bert,conneau2019cross,conneau2020unsupervised} made it possible to train the multilingual representations from monolingual data alone, relieving the need for parallel corpora.

In present days, the state of the art of cross-lingual text classification is held by multilingual LMs such as mBERT \cite{devlin-etal-2019-bert}, mBART \cite{liu2020multilingual}, XLM-R \cite{conneau2019cross}, and mT5 \cite{xue2021mt5}. Cross-lingual benchmarks such as XTREME \cite{hu2020xtreme} and XGLUE \cite{liang2020xglue} have helped established a reference for text classification and other NLP tasks over many languages. Generally, the models that are the highest in the leaderboards have been trained with masked language models, \textit{ELECTRA}-style models that use additional discriminative pretraining tasks \cite{chi2022xlme}, and models with cross-attention modules that build language interdependencies explicitly \cite{luo2021veco}, all typically using hundreds of  gigabytes or terabytes of pretraining data (e.g., 2.5 TB of Common Crawl data for XLM-R).

However, such large, pretrained multilingual models still suffer from significant performance limitations, particularly in low-resource languages. For example, \citet{ogueji2021small} have shown that a basic transformer directly trained in 11 low-resource African languages has neatly outperformed larger models such as mBERT and XLM-R in text classification,
even in languages present in the larger models. Other work has addressed the curse of multilinguality \cite{pfeiffer2020mad,pfeiffer2022lifting} by adding language-specific adapter modules \cite{houlsby2019parameter} or reparametrizing the transformer around the highest-resource language \cite{Artetxe2020}. However, significant performance differences between languages still persist.

The extensive availability of resources for machine translation such as datasets and pretrained multilingual MT models (e.g., many-to-many, many-to-one) justifies its continued exploration for cross-lingual text classification \cite{huang2019unicoder, conneau2020unsupervised, yu2022translate}. A well-practised approach is \textit{translate-and-train}, which first translates all the classification training data from the high-resource language (e.g., English) to the target language using an existing machine translation model, and then trains a classifier in the target language. In a sense, it could be objected that such an approach is not really ``cross-lingual'', in that the classification is performed by a model directly trained in the target language, making it more akin to a monolingual classifier trained with silver standard corpora. However, its main limitation lies in its inflexibility, since performing $T$ classification tasks over $L$ languages would require training $TL$ classification models. A more flexible alternative is offered by \textit{translate-and-test}, where the text from the target language is first translated to the high-resource language, and then classified using a classifier trained in the high-resource language. By comparison, this approach is genuinely cross-lingual and only requires training $T$ classification models in total. However, both approaches have proved able to achieve competitive results, provided that the machine translation step is sufficiently accurate.

Our work follows the translate-and-test approach, aiming to jointly fine-tune the machine translator and the classifier over few shots of the target classification task. 
This is in a similar vein with other works \cite{tebbifakhr2019machine, ponti2021modelling} which have fine-tuned a machine translator with reinforcement learning using the output of the downstream task classifier as reward. In this way, the machine translator can learn to adjust its translations optimally for the classification in the target language.
However, where available, differentiable objectives are generally preferable to reinforcement learning due to the high variance and instability of the latter. For this reason, in our approach, we leverage \textit{soft} translations (i.e., predictions of differentiable token embeddings rather than tokens) from the machine translation module to retain differentiability through the entire network and fine-tune the classifier and the translator jointly end-to-end.


\section{T3L}
\label{sec:t3l}

In this section, we formally describe the proposed approach, T3L, which essentially consists of two language models -- a machine translation language model (MT-LM) and a text classification language model (TC-LM) -- connected by a sequence of expected embeddings.

The MT-LM takes a sequence of tokens in the target language, $\{ x_1, \dots, x_n\}$, as input, and translates it into a sequence of tokens, $\{ t_1, \dots, t_m\}$, in the high-resource language (i.e., a language that has enough resources to build a strong classifier). At each decoding step $j$, given the input sequence and the previous decoded token, $t_{j-1}$, the LM generates a vector $\textbf{p}_j$ that represents the probability distribution over the words in the vocabulary, $V$:

\begin{equation}
\label{eq:mt_lm}
\textbf{p}_j = MT \mhyphen LM(t_{j-1}, x_1, \dots, x_n, \theta)
\end{equation}

\noindent where the learnable parameters of the MT-LM have been noted as $\theta$. Then, greedy decoding can be applied to generate the next translation token: 

\begin{equation}
\label{eq:greedy_argmax}
t_j = \argmax_{v = 1 \ldots V} p^v_j
\end{equation}

\noindent with $p^v_j$ the probability of the $v$-th token in the vocabulary, and $V$ the vocabulary size.

To generate the probability distribution vector, $\textbf{p}_j$, we use the standard softmax, but other approaches could be explored in alternative, including sparsemax \cite{martins2016softmax} and the Gumbel-softmax \cite{jang2017categorical}, to respectively encourage sparsity or controllable diversity in the probability vector. In addition, different decoding strategies such as beam search and sampling can be employed.

As the next step, the generated predictions are passed in input to the TC-LM model, yet ensuring that the whole network remains differentiable end-to-end. To this aim, the sequence of probability vectors, $\{ \textbf{p}_1, \dots, \textbf{p}_m\}$, generated by the MT-LM is used to ``mix'' the token embedding layer,  $\textbf{E}$, of the TC-LM to obtain a sequence of expected embeddings, $\{\bar{\textbf{e}}_1 \ldots \bar{\textbf{e}}_j \ldots \bar{\textbf{e}}_m\}$:

\begin{equation}
\label{eq:expected_emb}
\bar{\textbf{e}}_j = \mathbb{E}[\textbf{E}]_{\textbf{p}_j} = \sum_{v=1}^V p^v_j \textbf{e}^v
\end{equation}

\noindent where $\textbf{e}^v$ is the $v$-th embedding in the embedding layer of the TC-LM model, $\textbf{E}$. The expected embeddings are directly provided as input to the TC-LM, bypassing its embedding layer.

Finally, the TC-LM classifier predicts a text class given the sequence of expected embeddings:

\begin{equation}
\label{eq:dt_lm}
y = TC \mhyphen LM(\bar{\textbf{e}}_1, \dots, \bar{\textbf{e}}_m, \sigma)
\end{equation}

\noindent where the learnable parameters of the TC-LM have been noted as $\sigma$. Since the soft predictions from the MT-LM do not interrupt the backpropagation chain, the entire network can be trained end-to-end using common text classification training objectives. We use the cross-entropy loss for multi-class classification tasks and the binary cross-entropy loss on each label for multi-label classification tasks.

\begin{table}[t]	
	\begin{center}
		\centering
		\resizebox{0.45\textwidth}{!}{\begin{tabular}{|l|l|c|}
			\hline
			\textbf{Dataset}&\textbf{Languages}&\textbf{\# of labels}\\
			\hline
			XNLI&\begin{tabular}{@{}c@{}}ar,bg,de,el,es,fr,hi, \\ ru,sw,th,tr,ur,vi,zh\end{tabular}&3\\
			\hline
			MLDoc&de,es,fr,it,ja,ru,zh&4\\
			\hline
			MultiEURLEX&bg,el,nl,pl,pt,sl&21\\
			\hline
		\end{tabular}}
		\caption{Cross-lingual text classification datasets used for the experiments (NB: for MultiEURLEX, only a subset of the available languages).}
	   \label{tab:cross_lingual_datasets}
    \end{center}
    
\end{table}

%

\subsection{Vocabulary constraint}
\label{ssec:vocab_constraint}

Equation \ref{eq:expected_emb} generates an expected embedding by multiplying each token embedding in the embedding layer of the TC-LM model by the probability mass assigned to that same token by the MT-LM model. This implies that the vocabularies of the two LMs have to be aligned so that the same index in the two vocabularies refers to the same token.
In this paper, we address this by simply forcing the two LMs to share the same vocabulary, but a viable alternative would be to align the two vocabularies in a many-to-many manner in the style of optimal transport \cite{Xu2021}. We leave this to future exploration.

\subsection{Training strategy}

Given that T3L combines two LMs, the number of trainable parameters (i.e., $|\theta| + |\sigma|$) is approximately double that of a multilingual LM. This could become an issue if the available memory is not large enough to accommodate the fine-tuning of all the parameters. As a remediation, during fine-tuning we simply freeze some of the layers of both models. Specifically, we freeze the earlier layers of the MT-LM (e.g., layers closer to the input) and the later layers of the TC-LM (e.g., layers closer to the output), so as to concentrate the trainable parameters at the coupling of the two models. In this way, the number of trainable parameters becomes comparable to that of a multilingual LM. 


As in cross-lingual transfer learning, the TC-LM model is initialized with a multilingual LM and trained for the classification task in a high-resource language (e.g., English). In turn, the MT-LM model can be initialized with an off-the-shelf, pretrained multilingual MT model. However, if a pretrained translator is not available for the desired target language, any parallel corpus from the target language to the high-resource one can be used to train the MT-LM.

\begin{table*}[ht]
    \begin{center}
	    \begin{subtable}[h]{0.95\textwidth}
		\centering
		\resizebox{\textwidth}{!}{\begin{tabular}{|l|c|c|c|c|c|c|c|c|c|c|c|c|c|c|}
			\hline
			\multirow{2}{*}{\textbf{Training samples}}&\multicolumn{2}{c|}{ar}&\multicolumn{2}{c|}{bg}&\multicolumn{2}{c|}{de}&\multicolumn{2}{c|}{el}&\multicolumn{2}{c|}{es}&\multicolumn{2}{c|}{fr}&\multicolumn{2}{c|}{hi}\\
			\cline{2-15}
			&LM&T3L&LM&T3L&LM&T3L&LM&T3L&LM&T3L&LM&T3L&LM&T3L\\
			\hline
			Zero-shot&69.79&72.69&54.64&57.28&74.83&79.69&35.75&33.57&71.38&76.02&73.82&79.09&68.69&72.67\\
			Few-shot (10)&69.96&73.01&54.62&57.94&75.13&79.93&35.97&33.30&71.45&75.99&73.70&78.80&68.71&72.59\\
			Few-shot (100)&70.24&73.80&55.41&58.00&75.04&80.44&36.26&37.08&71.85&77.26&73.90&79.20&68.81&73.85\\
			\hline
		\end{tabular}}
		\end{subtable}
		\newline
        \vspace*{0.2 cm}
        \newline
		\begin{subtable}[h]{\textwidth}
		\centering
		\resizebox{\textwidth}{!}{\begin{tabular}{|l|c|c|c|c|c|c|c|c|c|c|c|c|c|c?c|c|}
			\hline
			\multirow{2}{*}{\textbf{Training samples}}&\multicolumn{2}{c|}{ru}&\multicolumn{2}{c|}{sw}&\multicolumn{2}{c|}{th}&\multicolumn{2}{c|}{tr}&\multicolumn{2}{c|}{ur}&\multicolumn{2}{c|}{vi}&\multicolumn{2}{c?}{zh}&\multicolumn{2}{c|}{\textbf{Average}}\\
			\cline{2-17}
			&LM&T3L&LM&T3L&LM&T3L&LM&T3L&LM&T3L&LM&T3L&LM&T3L&LM&T3L\\
			\hline
			Zero-shot&74.98&76.25&41.17&41.23&53.77&69.82&60.07&70.48&57.36&60.31&69.81&75.73&75.44&76.22&62.96&\textbf{67.43}\\
			Few-shot (10)&74.95&76.30&41.17&40.72&54.02&70.05&60.27&70.88&57.39&63.09&69.87&75.63&75.51&76.25&63.05&\textbf{67.46}\\
			Few-shot (100)&75.19&77.09&41.34&42.56&54.97&71.42&61.38&71.44&57.90&64.67&70.56&76.75&75.45&76.83&63.45&\textbf{68.60}\\
			\hline
		\end{tabular}}  
		\end{subtable}
		\caption{Classification accuracy over the XNLI test sets (average of three independent runs). }
	   \label{tab:XNLI_main_results}
	\end{center}
\end{table*}

\begin{table*}[ht]	
	\begin{center}
		\centering
		\resizebox{\textwidth}{!}{\begin{tabular}{|l|c|c|c|c|c|c|c|c|c|c|c|c|c|c?c|c|}
			\hline
			\multirow{2}{*}{\textbf{Training samples}}&\multicolumn{2}{c|}{de}&\multicolumn{2}{c|}{es}&\multicolumn{2}{c|}{fr}&\multicolumn{2}{c|}{it}&\multicolumn{2}{c|}{ja}&\multicolumn{2}{c|}{ru}&\multicolumn{2}{c?}{zh}&\multicolumn{2}{c|}{\textbf{Average}}\\
			\cline{2-17}
			&LM&T3L&LM&T3L&LM&T3L&LM&T3L&LM&T3L&LM&T3L&LM&T3L&LM&T3L\\
			\hline
			Zero-shot&77.62&90.14&69.05&73.24&75.82&87.28&56.11&72.43&68.47&62.12&49.53&65.66&70.89&79.44&66.78&\textbf{75.76}\\
			Few-shot (10)&79.20&92.39&69.07&77.49&76.24&89.04&56.28&73.21&68.56&65.50&50.61&67.42&72.12&82.07&67.44&\textbf{78.16}\\
			Few-shot (100)&89.79&92.85&82.01&83.05&86.05&89.68&72.43&77.51&75.83&74.24&80.76&76.10&84.66&84.37&81.65&\textbf{82.54}\\
			\hline
		\end{tabular}}
		\caption{Classification accuracy over the MLDoc test sets (average of three independent runs).}
	   \label{tab:MLDoc_main_results}
    \end{center}
    
\end{table*}

\begin{table*}[ht!]	
	\begin{center}
		\centering
		\resizebox{0.95\textwidth}{!}{\begin{tabular}{|l|c|c|c|c|c|c|c|c|c|c|c|c?c|c|}
			\hline
			\multirow{2}{*}{\textbf{Training samples}}&\multicolumn{2}{c|}{bg}&\multicolumn{2}{c|}{el}&\multicolumn{2}{c|}{nl}&\multicolumn{2}{c|}{pl}&\multicolumn{2}{c|}{pt}&\multicolumn{2}{c?}{sl}&\multicolumn{2}{c|}{\textbf{Average}}\\
			\cline{2-15}
			&LM&T3L&LM&T3L&LM&T3L&LM&T3L&LM&T3L&LM&T3L&LM&T3L\\
			\hline
			Zero-shot&65.41&75.46&37.00&33.62&74.69&82.84&77.66&83.67&73.03&83.17&69.69&80.47&66.25&\textbf{73.20}\\
			Few-shot (10)&71.62&75.34&37.61&35.95&80.25&82.87&81.78&82.90&79.76&82.65&74.83&79.64&70.98&\textbf{73.23}\\
			Few-shot (100)&79.53&79.19&54.58&46.12&83.06&83.38&84.03&84.50&83.30&83.70&81.31&81.89&\textbf{77.63}&76.46\\
			\hline
		\end{tabular}}
		\caption{Mean R-Precision (mRP) (Manning et al., 2009) over the MultiEURLEX test sets (average of three independent runs).}
	   \label{tab:MultiEURLEX_main_results}
    \end{center}
    
\end{table*}


\section{Experimental set-up}
\label{sec:experiments}

\subsection{Datasets}
\label{subsec:datasets}

We have carried out various experiments on three popular cross-lingual text classification datasets, namely XNLI \cite{conneau2018XNLI}, a cross-lingual natural language inference dataset; MLDoc \cite{schwenk2018corpus}, a corpus for multilingual news article classification; and MultiEURLEX \cite{chalkidis2021MultiEURLEX}, a multilingual legal, multi-label document classification dataset. For MultiEURLEX, we have conducted the experiments with only 6 of its 23 languages due to their much larger size, yet trying to broadly represent the available language families (Germanic, Romance, Slavic and Hellenic). Table \ref{tab:cross_lingual_datasets} shows the languages and the number of classes in each dataset. As common in cross-lingual transfer learning, English has been used as the high-resource language, while the other languages have been used for evaluation. We have also carried out few-shot, cross-lingual fine-tuning using 10 and 100 samples held out from the validation set of each language\footnote{Datasets and splits used in this work are publicly accessible via the link in the GitHub repository.}. 



\subsection{Model training}
\label{subsec:model_training}

As the base multilingual pretrained LM, we have adopted mBART \cite{liu2020multilingual} for both the MT-LM and the TC-LM, and also for the multilingual LM baseline. The main reason for choosing mBART is that it is a full encoder-decoder architecture that suits well both machine translation and classification tasks. As version, we have used a model pretrained for \textit{x}-to-\textit{en} translation in 50 different languages\footnote{Hugging Face model: \texttt{facebook/mbart-large-\\50-many-to-one-mmt}} as we expected it to provide a strong base for both models, and also for the baseline. We note that not all the languages present in the datasets are covered by this model, for example, Greek and Bulgarian are not. For this reason, for tokenization of these languages we have used the token set of related languages in the pretrained model, namely Macedonian for Greek and Russian for Bulgarian. 
As training, the TC-LM model has been trained for each downstream task for 10 epochs using the training data available in English, and the checkpoint with the best accuracy (XNLI, MLDoc) or mean R-Precision (mRP) \cite{chalkidis2021MultiEURLEX} (MultiEURLEX) over the validation set has been selected for testing. The mRP is used as the reference metric for the multi-label MultiEURLEX dataset to avoid biasing the performance with the samples with more labels. To prevent that, the R-Precision measures the precision of the $R$ most-probable labels for a given sample, with $R$ the number of its true positive labels (e.g., if a sample has 7 positive labels, the 7 most probable labels in the prediction are used to measure the precision). In this way, samples with different numbers of true labels span the same range in the metric. In turn, the mRP is just the average of the R-Precision over the entire sample set. All the models have been implemented in PyTorch using PyTorch Lightning\footnote{Code repository: \href{https://github.com/inigo-jauregi/t3l}{https://github.com/inigo-jauregi/t3l}}.

For the main experiments, we have used the trained models in a combination of zero-shot and few-shot configurations. In the zero-shot configuration, the trained model has been used as is. In the few-shot configuration, we have fine-tuned the pipeline in the target language with 10 and 100 samples, respectively, for each classification task. For performance comparison, we have used the same mBART model in a more conventional cross-lingual classification configuration (i.e., the target language as input and the predicted label as output), trained with the same English data for each classification task. In addition, we report performance comparisons with two other popular cross-lingual LMs, mBERT~\cite{devlin-etal-2019-bert} and XLM-R~\cite{conneau2019cross}. More details on the training, fine-tuning and hyperparameters are presented in the Appendix.


\section{Results}
\label{sec:results}

Table~\ref{tab:XNLI_main_results} shows the classification accuracy over the test sets of the XNLI dataset. The first remark is that the mBART baseline (noted simply as LM in the table) without any fine-tuning has reported a very different accuracy over the various languages, from a minimum of 35.75 percentage points (pp) for Greek to a maximum of 75.44 pp for Chinese, with a gap of nearly 40 pp. Conversely, T3L zero-shot has achieved a better accuracy in all languages but Greek, leading to an average accuracy improvement of +4.47 pp (62.96 vs 67.43), and impressive improvements in some cases (+16.05 pp in Thai and +10.41 in Turkish). As to be expected, both models have achieved the lowest accuracies on untrained languages (Greek, Bulgarian and Swahili \cite{liu2020multilingual}).
Few-shot fine-tuning in the target languages has led to an improvement across the board for both models, yet with T3L reporting an average improvement over mBART of +4.41 pp with 10 samples and +5.15 pp with 100.

Table~\ref{tab:MLDoc_main_results} shows the classification accuracy over the test sets of the MLDoc dataset. The zero-shot results show a similar trend to that of XNLI, with T3L reporting an average accuracy of 75.76 versus mBART's 66.78 (+8.98 pp). A similar trend also applies to the 10-shot fine-tuning, but the 100-shot fine-tuning has brought the average accuracy of the two models much closer together, with a difference of only 0.89 pp (82.54 for T3L vs 81.65 for mBART). This shows that for this dataset a more extensive fine-tuning has been able to realign the otherwise different performance of the two models. In general, the few-shot fine-tuning has improved the accuracy with all the languages, and quite dramatically so in some cases (for instance, an increase of 31.23 pp for Russian with mBART and 100 samples). For this dataset, mBART has performed better than the proposed approach in a few cases, namely Japanese, Russian and Chinese (100 samples). However, T3L has reported a higher accuracy in 16 configurations out of 21. 

Finally, Table~\ref{tab:MultiEURLEX_main_results} shows the mRP over the test sets of the MultiEURLEX dataset. For this dataset, T3L has achieved a higher average performance than mBART with zero shots (73.20 vs 66.25, i.e. +6.95 pp) and for 10 shot tuning (73.23 vs 70.98, i.e. +2.25 pp), but not for the 100-shot tuning (76.46 vs 77.63, i.e. -1.17 pp). However, it must be noted that the average mRP of T3L has been severely affected by its poor performance in Greek, a language not covered by the pretrained MT. T3L has still achieved a higher score in 14 out of 18 configurations.

In general, these results show that the proposed approach, based on the translate-and-test pipeline and soft translations, has been able to improve the performance over the LM baseline for many languages even in the zero-shot configuration. In addition, the end-to-end fine-tuning has helped further increase the accuracy in all tested cases.

\begin{figure}[t]
	\centering
	\includegraphics[width=0.95\linewidth]{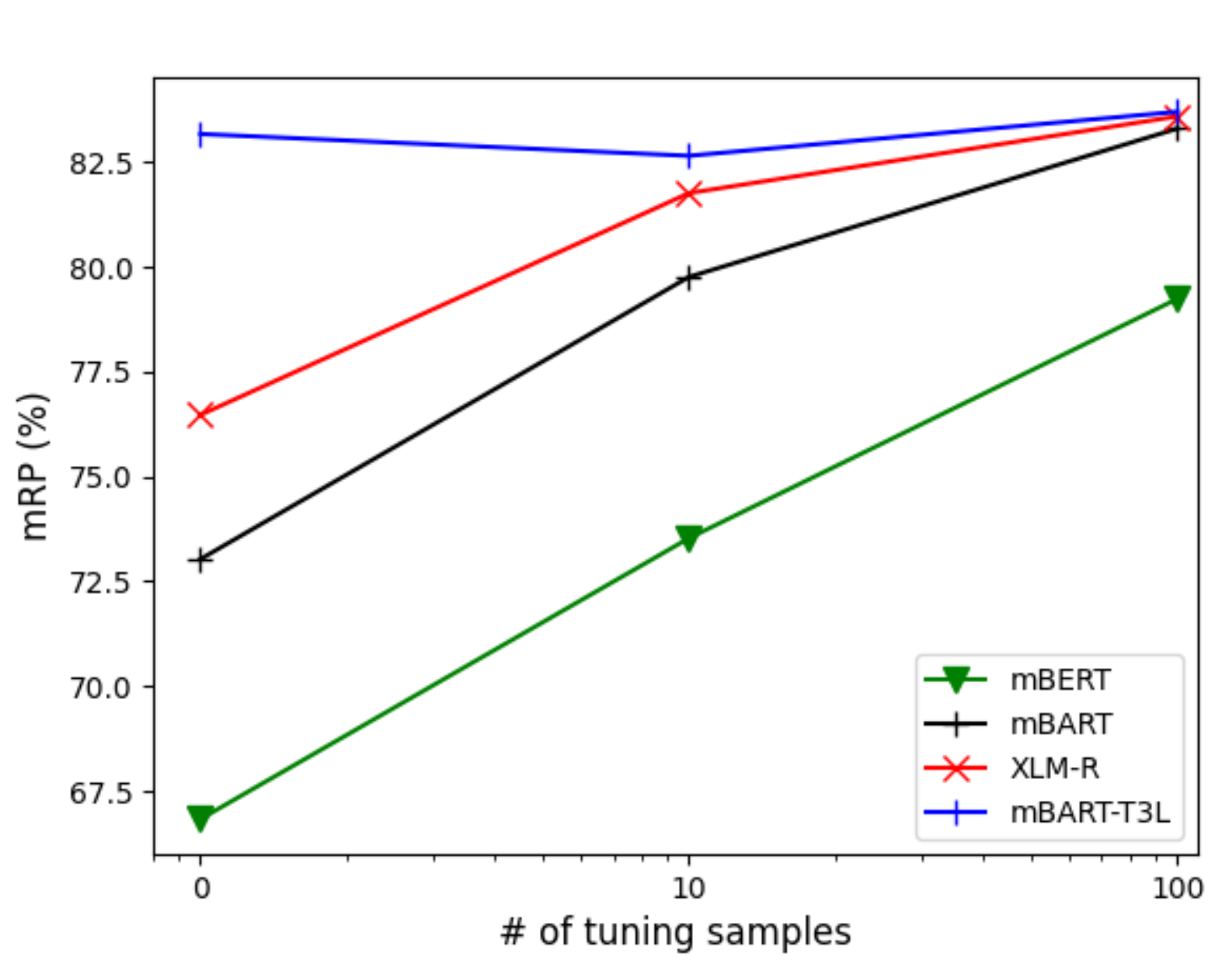}
	\caption{mRP (\%) of mBERT, XLM-R, mBART and mBART-T3L over the Portuguese (pt) MultiEURLEX test set. The values are an average of three independent runs.}
	\label{fig:multieurlex_pt}
\end{figure}

\begin{figure}[t]
	\centering
	\includegraphics[width=0.95\linewidth]{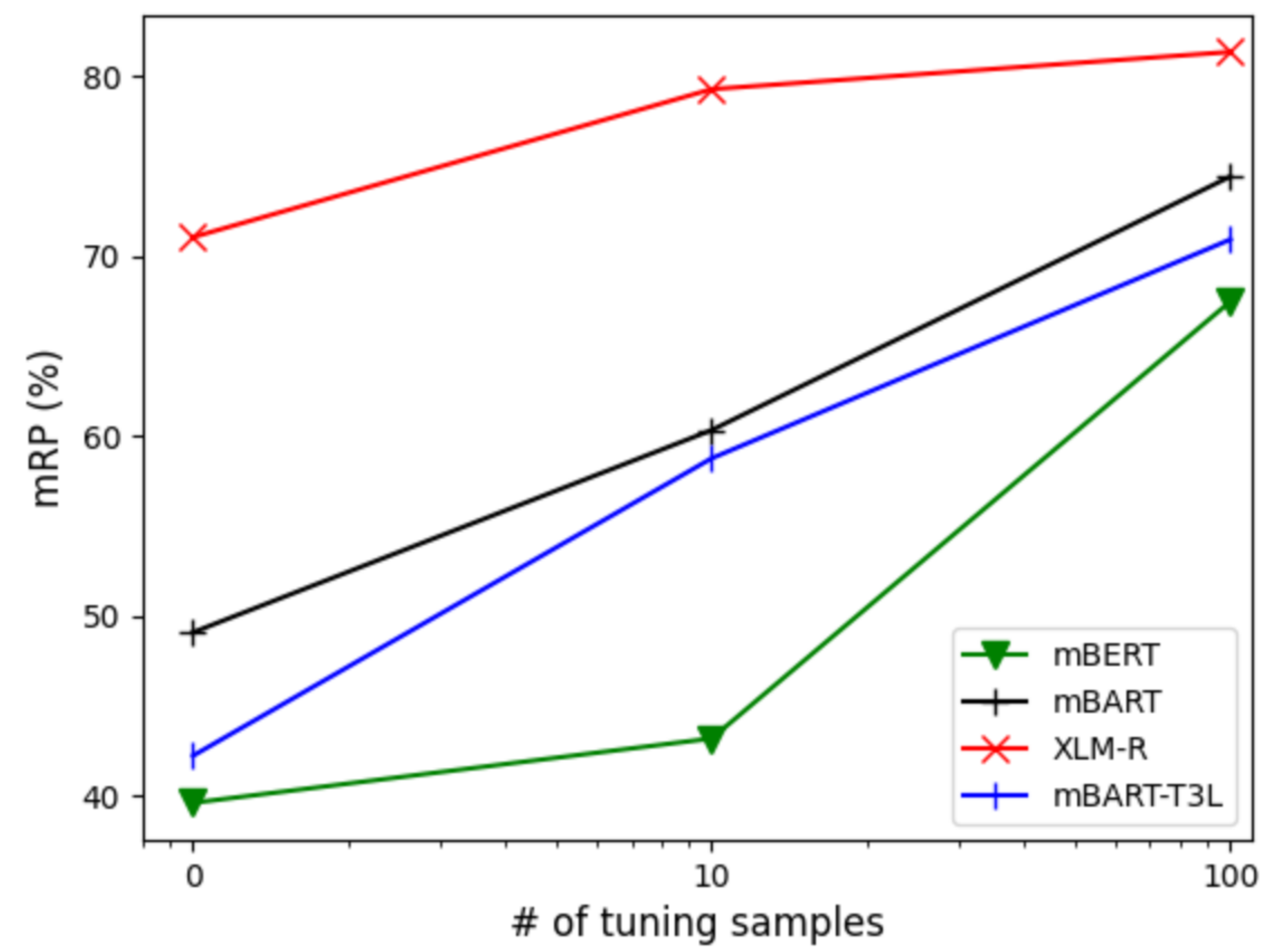}
	\caption{mRP (\%) of mBERT, XLM-R, mBART and mBART-T3L over the Greek (el) MultiEURLEX test set. The values are an average of three independent runs.}
	\label{fig:multieurlex_el}
\end{figure}


\subsection{Impact of dedicated translation training}
\label{ssec:translation_training}

\begin{table}[t]	
	\begin{center}
		\centering
		\resizebox{0.46\textwidth}{!}{\begin{tabular}{|l|l|c|c|c|c|}
			\hline
			\textbf{Languages}&\textbf{train}&\textbf{dev}&\textbf{test}&\textbf{Pr. BLEU}&\textbf{Tn. BLEU}\\
			\hline
			bg-en&109,616&1,000&5,000&16.18&46.00\\
			\hline
			el-en&30,582&1,000&5,000&0.32&38.61\\
			\hline
			sw-en&7,633&2,000&2,000&3.53&26.36\\
			\hline
		\end{tabular}}
		\caption{TED Talks translation datasets used to train the MT-LM model. The test-set BLEU scores of the pretrained mBART model (Pr.) and after dedicated training (Tn.) are reported to give a general idea of the translation quality.}
	   \label{tab:iwslt_data}
    \end{center}
\end{table}

\begin{table}[t]	
	\begin{center}
		\centering
		\resizebox{0.40\textwidth}{!}{\begin{tabular}{|l|l|c|c|c|c|}
			\hline
			\multirow{2}{*}{\textbf{Model}}&\multicolumn{3}{c|}{\textbf{XNLI}}&\multicolumn{2}{c|}{\textbf{MultiEURLEX}}\\
			\cline{2-6}
			&bg&el&sw&bg&el\\
			\hline
			T3L&78.04&70.03&59.68&80.80&41.91\\
			T3L (10)&78.29&70.42&60.50&81.77&58.77\\
			T3L (100)&\textbf{78.46}&\textbf{72.65}&\textbf{62.11}&\textbf{82.97}&70.94\\
			\hline
			LM&69.40&57.13&50.26&72.36&49.09\\
			LM (10)&71.40&60.42&51.36&79.17&60.34\\
			LM (100)&71.89&61.93&53.56&82.66&\textbf{74.44}\\
			\hline
		\end{tabular}}
		\caption{Accuracy (XNLI) and mRP (MultiEURLEX) with the improved MT-LM models.}
	   \label{tab:translation_training}
    \end{center}
\end{table}

T3L has achieved its lowest performance in Bulgarian, Greek and Swahili which are languages not covered by the pretrained mBART model. Therefore, in this section we explore whether a dedicated translation training of the MT-LM model may be able to improve the results for these languages on the downstream tasks. For training the MT-LM we have used the public parallel corpora of TED talks \cite{reimers2020}\footnote{\href{http://opus.nlpl.eu/TED2020-v1.php}{http://opus.nlpl.eu/TED2020-v1.php}}. Table \ref{tab:iwslt_data} shows our training, validation and test set splits, with the training data ranging from $\sim$7K to $\sim$100K parallel sentences\footnote{MT data splits are also publicly available via the link in the GitHub repository.} (overall, small sizes by contemporary standards). Three MT-LM models (one for each language pair) have been further trained for 10 epochs starting from the pretrained mBART model, and the checkpoint with the best validation BLEU retained. Table \ref{tab:iwslt_data} shows that the test-set BLEU scores were initially appalling, but have increased remarkably thanks to the dedicated training.

Table \ref{tab:translation_training} reports the results for the targeted languages with the improved translators. For a fair comparison with the cross-lingual mBART, we have also trained it with the same parallel corpus prior to the task learning, and included its results in Table \ref{tab:translation_training}. T3L has clearly improved the results in all the targeted languages, and outperformed the the LM baseline in four cases out of five. The improvements in Greek  have been particularly noticeable, with an accuracy increase of +35.57 pp with 100 shots compared to the model in Table \ref{tab:XNLI_main_results}, and also in Swahili, with an increase of +19.55 pp. The baseline LM has also improved, yet not to the same extent since the dedicated translation tuning has increased the performance gap between the models in four cases out of five.
For example, with 100-shot fine-tuning, the performance gap over the XNLI \textit{bg} test set has increased by +3.98 pp (from 58.00 $-$ 55.41 = +2.59 pp without the dedicated translation training to 78.46 $-$ 71.89 = +6.57 pp with translation training). Similarly, the performance gap has increased by +9.90 pp for XNLI el, +7.33 pp for XNLI sw, and +0.65 pp for MultiEURLEX bg. These results suggests that, at least for some languages, it may be more effective to improve the translator with a relatively small parallel corpus rather than using the same data to further train the baseline LM.


\subsection{Pretrained model selection}
\label{ssec:pretrained_model}

All the above results are direct comparisons between the proposed approach and the mBART baseline; yet, several other multilingual, encoder-only LMs are available for cross-lingual classification. To explore them, in Figures \ref{fig:multieurlex_pt} and \ref{fig:multieurlex_el} we report results for selected languages also including mBERT and XLM-R. Figure  \ref{fig:multieurlex_pt} shows the mRP over the Portuguese test set of MultiEURLEX. For this language, the proposed approach (noted as mBART-T3L for clarity) has achieved the highest mRP for both zero and few shots. XLM-R and mBART have instead achieved a comparable performance with 100 shots, while mBERT has been generally less accurate. In brief, one could say that mBART-T3L has offset the need for fine-tuning, most likely thanks to the fact that the Portuguese-to-English translation of the underlying mBART model is very accurate (we have measured it on the IWSLT 2014 test set, reporting a 61.6 BLEU score). In turn, Figure  \ref{fig:multieurlex_el} shows the mRP for Greek. For this language, XLM-R has achieved the highest mRP for both zero and few shots. 
The more modest performance of mBART-T3L can likely be explained by the fact that Greek is not part of the mBART pretrained languages, and the performance of its MT-LM Greek-to-English translation has proved less accurate even after the dedicated training (38.61 BLEU score on the IWSLT 2014 test set). Conversely, for this language the pretrained cross-lingual embeddings of XLM-R and the other LMs may have benefited from the many tokens shared with other close languages (e.g., Russian). In all cases, the performance gap between XML-R and all the other models has decreased substantially after fine-tuning.





\begin{table*}[ht!]	
	\begin{center}
		\centering
		\resizebox{0.95\textwidth}{!}{\begin{tabular}{|l|l|c|}
			\hline
			\multicolumn{2}{|l|}{\textbf{Original sentence}}&\textbf{Label}\\
			\hline
			src (es)&\begin{tabular}{@{}l@{}}\textbf{Premise}: Una vez detenido, KSM niega que Al Qaeda tuviera agentes en el sur de California. \\ \textbf{Hypothesis}: Al Qaeda puede no haber tenido agentes en California.\end{tabular}&entailment\\
			\hline
			\textbf{Model}&\textbf{Intermediate Translation}&\textbf{Pred}\\
			\hline
			\begin{tabular}{@{}l@{}}T3L\\(zero-shot)\end{tabular}&\begin{tabular}{@{}l@{}}\textbf{Premise}: Once arrested, KSM denies that Al Qaeda had \textcolor{blue}{operatives} in Southern California.\\\textbf{Hypothesis}: Al Qaeda may not have had agents in California.\end{tabular}&neutral\\
			\hline
			\begin{tabular}{@{}l@{}}T3L\\(few-shot 100)\end{tabular}&\begin{tabular}{@{}l@{}}\textbf{Premise}:  Once arrested, KSM denies that Al Qaeda had \textcolor{OliveGreen}{agents} in Southern California.\\\textbf{Hypothesis}: Al Qaeda may not have had agents in California.\end{tabular}&entailment\\
			\hline
		\end{tabular}}
		\caption{Qualitative example for XNLI in Spanish.}
	   \label{tab:XNLI_example1}
    \end{center}
    
\end{table*}

\begin{table*}[ht!]	
	\begin{center}
		\centering
		\resizebox{0.95\textwidth}{!}{\begin{tabular}{|l|l|c|}
			\hline
			\multicolumn{2}{|l|}{\textbf{Original sentence}}&\textbf{Label}\\
			\hline
			src (sw)&\begin{tabular}{@{}l@{}}\textbf{Premise}: Mwisho wa 1962, nilipata maelezo niende Washington, D.C.\\ \textbf{Hypothesis}: Niliambiwa kwenda DC.\end{tabular}&entailment\\
			\hline
			\textbf{Model}&\textbf{Intermediate Translation}&\textbf{Pred}\\
			\hline
			\begin{tabular}{@{}l@{}}T3L\\(zero-shot)\end{tabular}&\begin{tabular}{@{}l@{}}\textbf{Premise}: Mwisho wa 1962, melipata details niende Washington, D.C.\\\textbf{Hypothesis}: Niliambiwa leaves DC.\end{tabular}&contradiction\\
			\hline
			\begin{tabular}{@{}l@{}}T3L - MT trained\\(zero-shot)\end{tabular}&\begin{tabular}{@{}l@{}}\textbf{Premise}: In late 1962, I got some information \textcolor{red}{from somewhere in} Washington, D.C.\\\textbf{Hypothesis}: I was sent to DC.\end{tabular}&neutral\\
			\hline
			\begin{tabular}{@{}l@{}}T3L - MT trained\\(few-shot 100)\end{tabular}&\begin{tabular}{@{}l@{}}\textbf{Premise}:  In late 1962, I got some information \textcolor{OliveGreen}{about going to} Washington, D.C.\\\textbf{Hypothesis}: I was sent to DC.\end{tabular}&entailment\\
			\hline
		\end{tabular}}
		\caption{Qualitative example for XNLI in Swahili.}
	   \label{tab:XNLI_example2}
    \end{center}
    
\end{table*}


\subsection{Qualitative analysis of the intermediate translations}

Tables \ref{tab:XNLI_example1} and \ref{tab:XNLI_example2} show qualitative examples of the impact of the MT-LM training and the overall fine-tuning on the translations themselves. Note that the English translations shown in the tables are the ``argmaxed'' tokens ($t_j$ in Equation \ref{eq:greedy_argmax}) obtained from the sequence of probability distributions $\textbf{p}_j$ generated by the MT-LM model (Equation \ref{eq:mt_lm}). While in T3L the actual input to the TC-LM is the sequence of expected embeddings described in Equation \ref{eq:expected_emb}, the argmaxed tokens help illustrate the changes occurred in the translations after few-shot tuning.

Table \ref{tab:XNLI_example1} shows an example in Spanish from XNLI. In this case, all the translations are approximately comparable, but the TC-LM module has failed to correctly classify in the first two cases, likely because it has not been able to bridge the semantic gap between ``operatives'' and ``agents'' (or, more precisely, their soft translation counterparts). With the 100-shot fine-tuning, the translation has been somehow ``simplified'' to make inference easier, and the model has been able to predict the correct label. Such a functional simplification of the translation for the downstream task had also been reported by \citet{tebbifakhr2020machine}.
In turn, Table \ref{tab:XNLI_example2} shows an example in Swahili. 
The example shows that the 100-shot fine-tuning has been able to recover the correct label of the original Swahili text (\textit{entailment}). Without training, the MT-LM module has only translated a couple of words (``details'', ``leaves'').
Instead, with the MT-LM training the module has actually translated to English, and the model has been able to improve the classification to a less incorrect label. With the 100-shot fine-tuning, the changes in the translated tokens (from ``from somewhere in'' to ``about going to'') have certainly been pivotal to the correct classification.

For comparison, we have also tested the model using the actual argmaxed tokens as input to the TC-LM rather than the expected embeddings, but this resulted in lower accuracy in the downstream task. For example, over the Swahili XNLI test set the 100-shot T3L model with argmaxed translations has obtained an accuracy of 57.73 pp, compared to the 62.11 reported in Table \ref{tab:translation_training}. This shows that the overall improvement does not only stem from better translation, but also from the use of the soft predictions and end-to-end fine-tuning. For the specific example in the last row of Table \ref{tab:XNLI_example2}, the TC-LM with the argmaxed English translation has predicted \textit{neutral}, which is possibly an accurate classification of the translated text, but not an accurate prediction of the correct label of the Swahili text.


\subsection{Comparison with translate-and-train}
\label{subsec:translate_and_train}

The main advantage of the proposed approach with respect to translate-and-train is its flexibility, allowing reusing the same high-resource classifier for multiple languages. However, we
have also carried out some additional experiments to compare its performance with translate-and-train. To ensure that the models could be as comparable as possible, we have used the symmetric one-to-many mBART model\footnote{Hugging Face model: \\ \texttt{facebook/mbart-large-50-one-to-many-mmt}} that can translate from English to the same 50 languages and trained it with the same data. The classification training and validation sets have been translated from English to the target languages with this model, and then used for training dedicated classification models. For comparison, we have also carried out the same few-shot fine-tuning as for the proposed model.

Table \ref{tab:translate_and_train} shows the results for a language selection over the three tasks, showing that the relative performance of the two approaches varies with the language and the task. For instance, with the XNLI dataset translate-and-train has been more accurate than T3L for German, but less accurate for Greek. In contrast, with the MultiEURLEX dataset, T3L has performed worse than translate-and-train in Greek, yet better in Portuguese. We conclude that neither approach should be regarded as more accurate a priori, yet remark once more that translate-and-train is inherently more laborious due to its  requirement of separately translating the training data and re-training the classifier for each target language.  

\begin{table}[t]	
	\begin{center}
		\centering
		\resizebox{0.48\textwidth}{!}{\begin{tabular}{|l|c|c|c|c|c|c|}
			\hline
			\multirow{2}{*}{\textbf{Model}}&\multicolumn{2}{c|}{\textbf{XNLI}}&\multicolumn{2}{c|}{\textbf{MLDoc}}&\multicolumn{2}{c|}{\textbf{MultiEURLEX}}\\
			\cline{2-7}
			&de&el&it&ja&el&pt\\
			\hline
			T3L&79.69&33.57&72.43&62.12&33.62&83.17\\
			T3L (10)&79.93&33.30&73.21&65.50&35.95&82.65\\
			T3L (100)&80.44&\textbf{37.08}&77.52&74.24&46.12&\textbf{83.70}\\
			\hline
			Trans-train&81.51&35.55&72.34&76.38&29.20&71.83\\
			Trans-train (10)&\textbf{82.18}&35.62&75.06&78.17&34.03&74.35\\
			Trans-train (100)&81.88&36.83&\textbf{80.39}&\textbf{81.38}&\textbf{51.33}&78.32\\
			\hline
		\end{tabular}}
		\caption{Comparison of T3L with the translate-and-train approach for a selection of languages from the three tasks. The table reports the accuracy for XNLI and MLDoc, and the mRP for MultiEURLEX (average of three independent runs).}
	   \label{tab:translate_and_train}
    \end{center}
    
\end{table}

\subsection{Machine translation sensitivity analysis}

The translation quality of the MT-LM model has a principled impact on the eventual text classification accuracy. Therefore, in this section we present a sensitivity analysis to the translation quality as proxied by the BLEU score. To this aim, we have saved intermediate checkpoints during the training of the MT-LM models described in Section \ref{ssec:translation_training}, measured their BLEU score over the test set, and used them to evaluate T3L in the zero- and few-shot configurations.

\begin{figure}[t]
	\centering
	\includegraphics[width=0.95\linewidth]{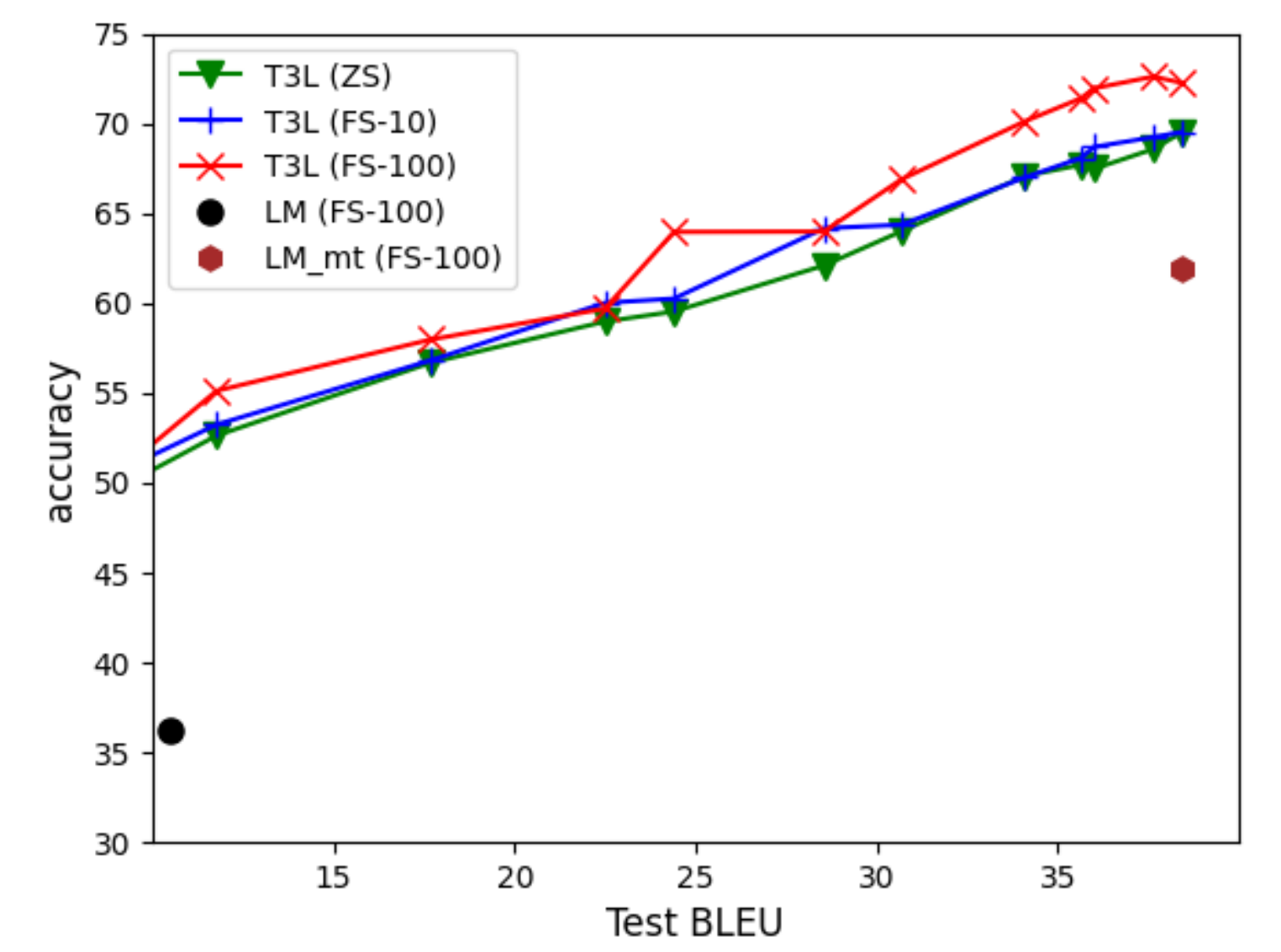}
	\caption{Sensitivity analysis of the impact of the MT-LM test BLEU score on the classification accuracy for the Greek XNLI test set. The MT-LM is an el-en model with dedicated training.}
	\label{fig:trans_sens_el_en_xnli}
\end{figure}

Figure \ref{fig:trans_sens_el_en_xnli} shows a plot of the accuracy against the BLEU score for the el-en MT-LM model on the XNLI test set. All the T3L models show a positive linear correlation between  the classification accuracy and the BLEU score. Since the accuracy does not show any sign of saturation, we speculate that it could increase further with even better translation quality. For comparison, the accuracy of the LM baseline has been much lower when using the original model
(35.75\%, Greek not included as language in LM pretraining), and has still been lower than that of T3L even when improved with the el-en dedicated training (61.93\%).

\begin{figure}[t]
	\centering
	\includegraphics[width=0.95\linewidth]{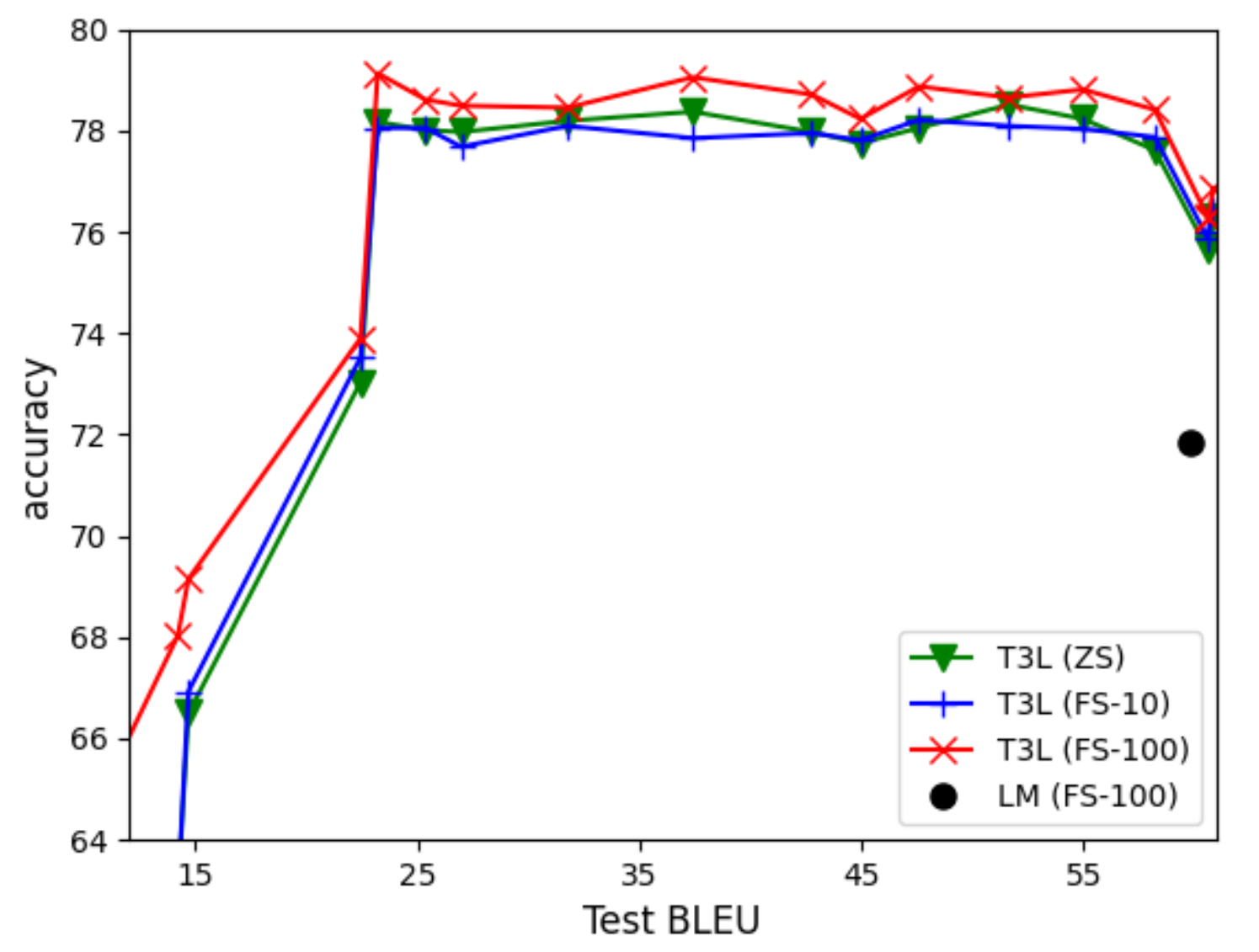}
	\caption{Sensitivity analysis of the impact of the MT-LM test BLEU score on the classification accuracy for the Spanish XNLI test set. The MT-LM is an es-en model with dedicated training.}
	\label{fig:trans_sens_es_en_xnli}
\end{figure}

In turn, Figure \ref{fig:trans_sens_es_en_xnli} shows an analogous plot for Spanish. In this case, an MT-LM model has been trained from scratch using the IWSLT 2014 TED Talks parallel data\footnote{\href{https://wit3.fbk.eu/2014-01}{https://wit3.fbk.eu/2014-01}} ($\sim$200K sentences) since the pretrained mBART already achieved a very high BLEU score on this language and could not support this exploration. For this language, the plot shows a saturation of the accuracy as the BLEU score increases, and even a small decline toward the end. 
It is noteworthy that the minimum BLEU score required to achieve the highest classification accuracies in this case has only been $\sim$22 pp, showing that in some cases a very high translation quality is not required.


\section{Discussion and Limitations}

Our results have shown that T3L is a competitive approach for cross-lingual text classification, particularly for low-resource languages. In broad terms, the proposed approach does not impose onerous training or inference requirements. The number of trained parameters is approximately the same as for a single language model. In all the experiments, we have been able to leverage an existing, pretrained translator without the need for dedicated training, with the exception of those languages that were not included in the pretraining corpora. The experiments have also shown that increasing the translation accuracy can have a positive impact on the classification accuracy, and that this can be achieved with parallel corpora of relatively small size. This seems a very attractive alternative to trying to improve the classification accuracy by increasing the already very large monolingual training corpora used for training the LM baseline. In addition, the same translation module can be shared by an unlimited number of different downstream tasks. It is also worth noting that an approach such as T3L is not confined to text classification tasks, but could instead be easily extended to perform natural language generation tasks.

Nevertheless, T3L also has its shortcomings. The main limitation that we identify is the vocabulary constraint described in Section \ref{ssec:vocab_constraint}. Since the output of the translator module is a sequence of embeddings rather than a detokenized string, the vocabulary of the classification module needs to be the same as that of the translator.
As such, we believe that it would be very useful to remove or mollify this constraint in future work, and for this we plan to explore differentiable alignment techniques such as optimal transport \cite{Xu2021}. This would eventually allow us to select completely independent models for the MT-LM and the TC-LM modules, making the choice more flexible.

\begin{table}[t]	
	\begin{center}
		\centering
		\resizebox{0.46\textwidth}{!}{\begin{tabular}{|l|c|c|c|}
			\hline
			\multirow{2}{*}{\textbf{Task}}&\textbf{LM}&\textbf{T3L}&Relative\\
            \cline{2-3}
            &ms/sample&ms/sample&slow-down\\
			\hline
            XNLI& 7 & 58 & $\times$8.28\\
			MLDoc& 25 & 298 & $\times$11.92\\
            MultiEURLEX& 1,427 & 1,586 & $\times$1.11\\
			\hline
		\end{tabular}}
		\caption{Per-sample inference times for a batch of 100 test samples for the baseline LM and T3L.}
	   \label{tab:inference_times}
    \end{center}
\end{table}

Another intrinsic limitation of translate-and-test approaches is their higher inference time,
mainly due to the sequential decoding introduced by the translation stage. To exemplify the relative slow-down, Table \ref{tab:inference_times} shows the time taken by the baseline LM and T3L to infer 100 test samples for each task, measured on an Intel Xeon Gold 6346 processor with 2 NVIDIA A40 GPUs. Note that the inference times vary substantially across these tasks due to their different average document length and the number of labels to be inferred (single for XNLI and MLDoc vs multiple for MultiEURLEX). The values show that T3L has proved moderately slower only over MultiEURLEX (1,586 ms vs 1,427 ms per sample), but it has been slower by an order of magnitude over XNLI and MLDoc.
However, we believe that this is not an impediment to the applicability of the proposed approach, given that the absolute inference times remain contained and they could be significantly sped-up with the use of parallel GPUs, model serialization and more efficient coding. In addition, in future work we will explore the use of parallel decoding with non-autoregressive transformers \cite{gu2018nat,qian2021glancing} which are catching up in performance with their non-autoregressive counterparts and may be able to further abate the overall inference latency.

\section{Conclusion}

This paper has presented T3L, a novel approach for cross-lingual text classification built upon the translate-and-test strategy and allowing end-to-end fine-tuning thanks to its use of soft translations.
Extensive experimental results over three benchmark cross-lingual datasets have shown that T3L is a performing approach that has worked particularly well over low-resource languages and has outperformed a completely comparable cross-lingual LM baseline in the vast majority of cases. While a model such as T3L certainly introduces some overheads compared to a conventional cross-lingual LM in terms of memory requirements and inference speed, we hope that its attractive classification performance may spark a renewed interest in the translate-and-test pipeline.

\bibliography{custom}

\begin{thebibliography}{39}
\expandafter\ifx\csname natexlab\endcsname\relax\def\natexlab#1{#1}\fi

\bibitem[{Artetxe et~al.(2020)Artetxe, Ruder, and Yogatama}]{Artetxe2020}
Mikel Artetxe, Sebastian Ruder, and Dani Yogatama. 2020.
\newblock On the cross-lingual transferability of monolingual representations.
\newblock In \emph{Proceedings of the 58th Annual Meeting of the Association
  for Computational Linguistics}, pages 4623--4637.

\bibitem[{Bel et~al.(2003)Bel, Koster, and Villegas}]{Bel2003}
N{\'{u}}ria Bel, Cornelis H.~A. Koster, and Marta Villegas. 2003.
\newblock Cross-lingual text categorization.
\newblock In \emph{Proceedings of the 7th European Conference on Digital
  Libraries}, volume 2769 of \emph{Lecture Notes in Computer Science}, pages
  126--139.

\bibitem[{Chalkidis et~al.(2021)Chalkidis, Fergadiotis, and
  Androutsopoulos}]{chalkidis2021MultiEURLEX}
Ilias Chalkidis, Manos Fergadiotis, and Ion Androutsopoulos. 2021.
\newblock {M}ulti{EURLEX} - a multi-lingual and multi-label legal document
  classification dataset for zero-shot cross-lingual transfer.
\newblock In \emph{Proceedings of the Conference on Empirical Methods in
  Natural Language Processing}, pages 6974--6996.

\bibitem[{Chi et~al.(2022)Chi, Huang, Dong, Ma, Zheng, Singhal, Bajaj, Song,
  Mao, Huang, and Wei}]{chi2022xlme}
Zewen Chi, Shaohan Huang, Li~Dong, Shuming Ma, Bo~Zheng, Saksham Singhal, Payal
  Bajaj, Xia Song, Xian-Ling Mao, Heyan Huang, and Furu Wei. 2022.
\newblock {XLM}-{E}: Cross-lingual language model pre-training via {ELECTRA}.
\newblock In \emph{Proceedings of the 60th Annual Meeting of the Association
  for Computational Linguistics}, pages 6170--6182.

\bibitem[{Conneau et~al.(2020)Conneau, Khandelwal, Goyal, Chaudhary, Wenzek,
  Guzm{\'a}n, Grave, Ott, Zettlemoyer, and Stoyanov}]{conneau2020unsupervised}
Alexis Conneau, Kartikay Khandelwal, Naman Goyal, Vishrav Chaudhary, Guillaume
  Wenzek, Francisco Guzm{\'a}n, Edouard Grave, Myle Ott, Luke Zettlemoyer, and
  Veselin Stoyanov. 2020.
\newblock Unsupervised cross-lingual representation learning at scale.
\newblock In \emph{Proceedings of the 58th Annual Meeting of the Association
  for Computational Linguistics}, pages 8440--8451.

\bibitem[{Conneau and Lample(2019)}]{conneau2019cross}
Alexis Conneau and Guillaume Lample. 2019.
\newblock Cross-lingual language model pretraining.
\newblock \emph{Advances in Neural Information Processing Systems}, 32.

\bibitem[{Conneau et~al.(2018)Conneau, Rinott, Lample, Williams, Bowman,
  Schwenk, and Stoyanov}]{conneau2018XNLI}
Alexis Conneau, Ruty Rinott, Guillaume Lample, Adina Williams, Samuel~R.
  Bowman, Holger Schwenk, and Veselin Stoyanov. 2018.
\newblock Xnli: Evaluating cross-lingual sentence representations.
\newblock In \emph{Proceedings of the Conference on Empirical Methods in
  Natural Language Processing}, pages 2475--2485.

\bibitem[{Devlin et~al.(2019)Devlin, Chang, Lee, and
  Toutanova}]{devlin-etal-2019-bert}
Jacob Devlin, Ming-Wei Chang, Kenton Lee, and Kristina Toutanova. 2019.
\newblock {BERT}: Pre-training of deep bidirectional transformers for language
  understanding.
\newblock In \emph{Proceedings of the 2019 Conference of the North {A}merican
  Chapter of the Association for Computational Linguistics: Human Language
  Technologies}, pages 4171--4186.

\bibitem[{Gu et~al.(2018)Gu, Bradbury, Xiong, Li, and Socher}]{gu2018nat}
Jiatao Gu, James Bradbury, Caiming Xiong, Victor~O.K. Li, and Richard Socher.
  2018.
\newblock Non-autoregressive neural machine translation.
\newblock In \emph{International Conference on Learning Representations}.

\bibitem[{Houlsby et~al.(2019)Houlsby, Giurgiu, Jastrzebski, Morrone,
  De~Laroussilhe, Gesmundo, Attariyan, and Gelly}]{houlsby2019parameter}
Neil Houlsby, Andrei Giurgiu, Stanislaw Jastrzebski, Bruna Morrone, Quentin
  De~Laroussilhe, Andrea Gesmundo, Mona Attariyan, and Sylvain Gelly. 2019.
\newblock Parameter-efficient transfer learning for {NLP}.
\newblock In \emph{Proceedings of the 36th International Conference on Machine
  Learning}, pages 2790--2799.

\bibitem[{Hu et~al.(2020)Hu, Ruder, Siddhant, Neubig, Firat, and
  Johnson}]{hu2020xtreme}
Junjie Hu, Sebastian Ruder, Aditya Siddhant, Graham Neubig, Orhan Firat, and
  Melvin Johnson. 2020.
\newblock {XTREME}: A massively multilingual multi-task benchmark for
  evaluating cross-lingual generalisation.
\newblock In \emph{Proceedings of the 37th International Conference on Machine
  Learning}, volume 119, pages 4411--4421.

\bibitem[{Huang et~al.(2019)Huang, Liang, Duan, Gong, Shou, Jiang, and
  Zhou}]{huang2019unicoder}
Haoyang Huang, Yaobo Liang, Nan Duan, Ming Gong, Linjun Shou, Daxin Jiang, and
  Ming Zhou. 2019.
\newblock {U}nicoder: A universal language encoder by pre-training with
  multiple cross-lingual tasks.
\newblock In \emph{Proceedings of the 2019 Conference on Empirical Methods in
  Natural Language Processing and the 9th International Joint Conference on
  Natural Language Processing}, pages 2485--2494.

\bibitem[{Jang et~al.(2017)Jang, Gu, and Poole}]{jang2017categorical}
Eric Jang, Shixiang Gu, and Ben Poole. 2017.
\newblock Categorical reparameterization with gumbel-softmax.
\newblock In \emph{Proceedings of the International Conference on Learning
  Representations}.

\bibitem[{Jauregi~Unanue et~al.(2021)Jauregi~Unanue, Parnell, and
  Piccardi}]{jauregi2021berttune}
Inigo Jauregi~Unanue, Jacob Parnell, and Massimo Piccardi. 2021.
\newblock {BERTT}une: Fine-tuning neural machine translation with {BERTS}core.
\newblock In \emph{Proceedings of the 59th Annual Meeting of the Association
  for Computational Linguistics and the 11th International Joint Conference on
  Natural Language Processing}, pages 915--924.

\bibitem[{Joshi et~al.(2020)Joshi, Santy, Budhiraja, Bali, and
  Choudhury}]{joshi2020state}
Pratik Joshi, Sebastin Santy, Amar Budhiraja, Kalika Bali, and Monojit
  Choudhury. 2020.
\newblock The state and fate of linguistic diversity and inclusion in the {NLP}
  world.
\newblock In \emph{Proceedings of the 58th Annual Meeting of the Association
  for Computational Linguistics}, pages 6282--6293.

\bibitem[{Klementiev et~al.(2012)Klementiev, Titov, and
  Bhattarai}]{Klementiev2012}
Alexandre Klementiev, Ivan Titov, and Binod Bhattarai. 2012.
\newblock Inducing crosslingual distributed representations of words.
\newblock In \emph{Proceedings of the 24th International Conference on
  Computational Linguistics}, pages 1459--1474.

\bibitem[{Kreutzer et~al.(2022)Kreutzer, Caswell, Wang, Wahab, van Esch,
  Ulzii-Orshikh, Tapo, Subramani, Sokolov, Sikasote, Setyawan, Sarin, Samb,
  Sagot, Rivera, Rios, Papadimitriou, Osei, Suarez, Orife, Ogueji, Rubungo,
  Nguyen, Müller, Müller, Muhammad, Muhammad, Mnyakeni, Mirzakhalov,
  Matangira, Leong, Lawson, Kudugunta, Jernite, Jenny, Firat, Dossou, Dlamini,
  de~Silva, Çabuk Ballı, Biderman, Battisti, Baruwa, Bapna, Baljekar, Azime,
  Awokoya, Ataman, Ahia, Ahia, Agrawal, and Adeyemi}]{kreutzer2022quality}
Julia Kreutzer, Isaac Caswell, Lisa Wang, Ahsan Wahab, Daan van Esch,
  Nasanbayar Ulzii-Orshikh, Allahsera Tapo, Nishant Subramani, Artem Sokolov,
  Claytone Sikasote, Monang Setyawan, Supheakmungkol Sarin, Sokhar Samb,
  Benoît Sagot, Clara Rivera, Annette Rios, Isabel Papadimitriou, Salomey
  Osei, Pedro~Ortiz Suarez, Iroro Orife, Kelechi Ogueji, Andre~Niyongabo
  Rubungo, Toan~Q. Nguyen, Mathias Müller, André Müller, Shamsuddeen~Hassan
  Muhammad, Nanda Muhammad, Ayanda Mnyakeni, Jamshidbek Mirzakhalov,
  Tapiwanashe Matangira, Colin Leong, Nze Lawson, Sneha Kudugunta, Yacine
  Jernite, Mathias Jenny, Orhan Firat, Bonaventure F.~P. Dossou, Sakhile
  Dlamini, Nisansa de~Silva, Sakine Çabuk Ballı, Stella Biderman, Alessia
  Battisti, Ahmed Baruwa, Ankur Bapna, Pallavi Baljekar, Israel~Abebe Azime,
  Ayodele Awokoya, Duygu Ataman, Orevaoghene Ahia, Oghenefego Ahia, Sweta
  Agrawal, and Mofetoluwa Adeyemi. 2022.
\newblock {Quality at a glance: An audit of web-crawled multilingual datasets}.
\newblock \emph{Transactions of the Association for Computational Linguistics},
  10:50--72.

\bibitem[{Lewis et~al.(2020)Lewis, Liu, Goyal, Ghazvininejad, Mohamed, Levy,
  Stoyanov, and Zettlemoyer}]{lewis-etal-2020-bart}
Mike Lewis, Yinhan Liu, Naman Goyal, Marjan Ghazvininejad, Abdelrahman Mohamed,
  Omer Levy, Veselin Stoyanov, and Luke Zettlemoyer. 2020.
\newblock {BART}: Denoising sequence-to-sequence pre-training for natural
  language generation, translation, and comprehension.
\newblock In \emph{Proceedings of the 58th Annual Meeting of the Association
  for Computational Linguistics}, pages 7871--7880.

\bibitem[{Liang et~al.(2020)Liang, Duan, Gong, Wu, Guo, Qi, Gong, Shou, Jiang,
  Cao, Fan, Zhang, Agrawal, Cui, Wei, Bharti, Qiao, Chen, Wu, Liu, Yang,
  Campos, Majumder, and Zhou}]{liang2020xglue}
Yaobo Liang, Nan Duan, Yeyun Gong, Ning Wu, Fenfei Guo, Weizhen Qi, Ming Gong,
  Linjun Shou, Daxin Jiang, Guihong Cao, Xiaodong Fan, Ruofei Zhang, Rahul
  Agrawal, Edward Cui, Sining Wei, Taroon Bharti, Ying Qiao, Jiun-Hung Chen,
  Winnie Wu, Shuguang Liu, Fan Yang, Daniel Campos, Rangan Majumder, and Ming
  Zhou. 2020.
\newblock {XGLUE}: A new benchmark dataset for cross-lingual pre-training,
  understanding and generation.
\newblock In \emph{Proceedings of the Conference on Empirical Methods in
  Natural Language Processing}, pages 6008--6018.

\bibitem[{Liu et~al.(2020)Liu, Gu, Goyal, Li, Edunov, Ghazvininejad, Lewis, and
  Zettlemoyer}]{liu2020multilingual}
Yinhan Liu, Jiatao Gu, Naman Goyal, Xian Li, Sergey Edunov, Marjan
  Ghazvininejad, Mike Lewis, and Luke Zettlemoyer. 2020.
\newblock Multilingual denoising pre-training for neural machine translation.
\newblock \emph{Transactions of the Association for Computational Linguistics},
  8:726--742.

\bibitem[{Luo et~al.(2021)Luo, Wang, Liu, Liu, Bi, Huang, Huang, and
  Si}]{luo2021veco}
Fuli Luo, Wei Wang, Jiahao Liu, Yijia Liu, Bin Bi, Songfang Huang, Fei Huang,
  and Luo Si. 2021.
\newblock {VECO}: Variable and flexible cross-lingual pre-training for language
  understanding and generation.
\newblock In \emph{Proceedings of the 59th Annual Meeting of the Association
  for Computational Linguistics and the 11th International Joint Conference on
  Natural Language Processing}, pages 3980--3994.

\bibitem[{Martins and Astudillo(2016)}]{martins2016softmax}
Andre Martins and Ramon Astudillo. 2016.
\newblock From softmax to sparsemax: {A} sparse model of attention and
  multi-label classification.
\newblock In \emph{International Conference on Machine Learning}, pages
  1614--1623.

\bibitem[{Ogueji et~al.(2021)Ogueji, Zhu, and Lin}]{ogueji2021small}
Kelechi Ogueji, Yuxin Zhu, and Jimmy Lin. 2021.
\newblock Small data? {N}o problem! {E}xploring the viability of pretrained
  multilingual language models for low-resourced languages.
\newblock In \emph{Proceedings of the 1st Workshop on Multilingual
  Representation Learning}, pages 116--126.

\bibitem[{Peters et~al.(2018)Peters, Neumann, Iyyer, Gardner, Clark, Lee, and
  Zettlemoyer}]{peters-etal-2018-deep}
Matthew~E. Peters, Mark Neumann, Mohit Iyyer, Matt Gardner, Christopher Clark,
  Kenton Lee, and Luke Zettlemoyer. 2018.
\newblock Deep contextualized word representations.
\newblock In \emph{Proceedings of the Conference of the North {A}merican
  Chapter of the Association for Computational Linguistics: Human Language
  Technologies}, pages 2227--2237.

\bibitem[{Pfeiffer et~al.(2022)Pfeiffer, Goyal, Lin, Li, Cross, Riedel, and
  Artetxe}]{pfeiffer2022lifting}
Jonas Pfeiffer, Naman Goyal, Xi~Victoria Lin, Xian Li, James Cross, Sebastian
  Riedel, and Mikel Artetxe. 2022.
\newblock Lifting the curse of multilinguality by pre-training modular
  transformers.
\newblock In \emph{Proceedings of the Conference of the North {A}merican
  Chapter of the Association for Computational Linguistics}, pages 3479--3495.

\bibitem[{Pfeiffer et~al.(2020)Pfeiffer, Vuli{\'c}, Gurevych, and
  Ruder}]{pfeiffer2020mad}
Jonas Pfeiffer, Ivan Vuli{\'c}, Iryna Gurevych, and Sebastian Ruder. 2020.
\newblock {MAD-X}: {A}n adapter-based framework for multi-task cross-lingual
  transfer.
\newblock In \emph{Proceedings of the Conference on Empirical Methods in
  Natural Language Processing}, pages 7654--7673.

\bibitem[{Ponti et~al.(2021)Ponti, Kreutzer, Vuli{\'c}, and
  Reddy}]{ponti2021modelling}
Edoardo~Maria Ponti, Julia Kreutzer, Ivan Vuli{\'c}, and Siva Reddy. 2021.
\newblock Modelling latent translations for cross-lingual transfer.
\newblock \emph{arXiv preprint arXiv:2107.11353}.

\bibitem[{Qian et~al.(2021)Qian, Zhou, Bao, Wang, Qiu, Zhang, Yu, and
  Li}]{qian2021glancing}
Lihua Qian, Hao Zhou, Yu~Bao, Mingxuan Wang, Lin Qiu, Weinan Zhang, Yong Yu,
  and Lei Li. 2021.
\newblock Glancing transformer for non-autoregressive neural machine
  translation.
\newblock In \emph{Proceedings of the 59th Annual Meeting of the Association
  for Computational Linguistics and the 11th International Joint Conference on
  Natural Language Processing}, pages 1993--2003.

\bibitem[{Radford et~al.(2018)Radford, Narasimhan, Salimans, and
  Sutskever}]{radford2018improving}
Alec Radford, Karthik Narasimhan, Tim Salimans, and Ilya Sutskever. 2018.
\newblock Improving language understanding by generative pre-training.
\newblock \emph{URL:
  \href{https://s3-us-west-2.amazonaws.com/openai-assets/research-covers/language-unsupervised/language_understanding_paper.pdf}{https://s3-us-west-2.amazonaws.com/openai-assets/research-covers/language-unsupervised/\\language\_understanding\_paper.pdf}}.

\bibitem[{Raffel et~al.(2020)Raffel, Shazeer, Roberts, Lee, Narang, Matena,
  Zhou, Li, and Liu}]{raffel2020t5}
Colin Raffel, Noam Shazeer, Adam Roberts, Katherine Lee, Sharan Narang, Michael
  Matena, Yanqi Zhou, Wei Li, and Peter~J. Liu. 2020.
\newblock Exploring the limits of transfer learning with a unified text-to-text
  transformer.
\newblock \emph{Journal of Machine Learning Research}, 21(140):1--67.

\bibitem[{Reimers and Gurevych(2020)}]{reimers2020}
Nils Reimers and Iryna Gurevych. 2020.
\newblock Making monolingual sentence embeddings multilingual using knowledge
  distillation.
\newblock In \emph{Proceedings of the Conference on Empirical Methods in
  Natural Language Processing}, pages 4512--4525.

\bibitem[{Schwenk and Li(2018)}]{schwenk2018corpus}
Holger Schwenk and Xian Li. 2018.
\newblock A corpus for multilingual document classification in eight languages.
\newblock In \emph{Proceedings of the Eleventh International Conference on
  Language Resources and Evaluation}.

\bibitem[{Tebbifakhr et~al.(2019)Tebbifakhr, Bentivogli, Negri, and
  Turchi}]{tebbifakhr2019machine}
Amirhossein Tebbifakhr, Luisa Bentivogli, Matteo Negri, and Marco Turchi. 2019.
\newblock Machine translation for machines: the sentiment classification use
  case.
\newblock In \emph{Proceedings of the Conference on Empirical Methods in
  Natural Language Processing and the 9th International Joint Conference on
  Natural Language Processing}, pages 1368--1374.

\bibitem[{Tebbifakhr et~al.(2020)Tebbifakhr, Negri, and
  Turchi}]{tebbifakhr2020machine}
Amirhossein Tebbifakhr, Matteo Negri, and Marco Turchi. 2020.
\newblock Machine-oriented {NMT} adaptation for zero-shot {NLP} tasks:
  Comparing the usefulness of close and distant languages.
\newblock In \emph{Proceedings of the 7th Workshop on NLP for Similar
  Languages, Varieties and Dialects}, pages 36--46.

\bibitem[{Vaswani et~al.(2017)Vaswani, Shazeer, Parmar, Uszkoreit, Jones,
  Gomez, Kaiser, and Polosukhin}]{vaswani2017attention}
Ashish Vaswani, Noam Shazeer, Niki Parmar, Jakob Uszkoreit, Llion Jones,
  Aidan~N Gomez, \L~ukasz Kaiser, and Illia Polosukhin. 2017.
\newblock Attention is all you need.
\newblock In \emph{Advances in Neural Information Processing Systems}.

\bibitem[{Wu et~al.(2008)Wu, Wang, and Lu}]{Wu2008}
Ke~Wu, Xiaolin Wang, and Bao{-}Liang Lu. 2008.
\newblock Cross language text categorization using a bilingual lexicon.
\newblock In \emph{Third International Joint Conference on Natural Language
  Processing}, pages 165--172.

\bibitem[{Xu et~al.(2021)Xu, Zhou, Gan, Zheng, and Li}]{Xu2021}
Jingjing Xu, Hao Zhou, Chun Gan, Zaixiang Zheng, and Lei Li. 2021.
\newblock Vocabulary learning via optimal transport for neural machine
  translation.
\newblock In \emph{Proceedings of the 59th Annual Meeting of the Association
  for Computational Linguistics and the 11th International Joint Conference on
  Natural Language Processing}, pages 7361--7373.

\bibitem[{Xue et~al.(2021)Xue, Constant, Roberts, Kale, Al-Rfou, Siddhant,
  Barua, and Raffel}]{xue2021mt5}
Linting Xue, Noah Constant, Adam Roberts, Mihir Kale, Rami Al-Rfou, Aditya
  Siddhant, Aditya Barua, and Colin Raffel. 2021.
\newblock m{T}5: A massively multilingual pre-trained text-to-text transformer.
\newblock In \emph{Proceedings of the Conference of the North American Chapter
  of the Association for Computational Linguistics: Human Language
  Technologies}, pages 483--498.

\bibitem[{Yu et~al.(2022)Yu, Sun, Zhang, and Jiang}]{yu2022translate}
Sicheng Yu, Qianru Sun, Hao Zhang, and Jing Jiang. 2022.
\newblock Translate-train embracing translationese artifacts.
\newblock In \emph{Proceedings of the 60th Annual Meeting of the Association
  for Computational Linguistics}, pages 362--370.

\end{thebibliography}
\bibliographystyle{acl_natbib}

\appendix

\newpage

\section{Appendix}
\label{sec:appendix}




Table \ref{tab:mt_lm_hyp} shows the hyperparameters employed for the training of the MT-LM and TC-LM models and the fine-tuning of the overall T3L model.

For a fair comparison of T3L with standard few-shot cross-lingual transfer learning, we have generated a robust baseline (named LM in Tables \ref{tab:XNLI_main_results}, \ref{tab:MLDoc_main_results} and \ref{tab:MultiEURLEX_main_results}) using the same TC-LM models described in Section \ref{subsec:model_training} and the same few-shot fine-tuning and validation data described in Section \ref{subsec:datasets}. All models have been trained for 10 epochs, and the checkpoints with the highest validation accuracy or mRP have been used for testing.

\begin{table}[ht!]	
	\begin{center}
		\centering
		\resizebox{0.45\textwidth}{!}{\begin{tabular}{|l|c|c|c|}
			\hline
			\textbf{Hyperparameter}&\textbf{MT-LM}&\textbf{TC-LM}&\textbf{T3L}\\
			\hline
			batch size&8&8&1\\
			\hline
			gradient accumulation&2&2&1\\
			\hline
			max gradient norm&1.0&1&1\\
			\hline
			learning rate&3e-5&3e-6&3e-6\\
			\hline
			warmup steps&500&500&0\\
			\hline
			weight decay&0.01&0.01&0.01\\
			\hline
			optimizer&AdamW&AdamW&AdamW\\
			\hline
			epochs&10&10&10\\
			\hline
			max input sequence length&170&170/512$^*$&85/256$^*$\\
			\hline
			max output sequence length&170&170/512$^*$&85/256$^*$\\
			
			\hline
		\end{tabular}}
		\caption{Hyperparameter selection for the MT-LM, TC-LM and T3L models.}
	   \label{tab:mt_lm_hyp}
    \end{center}
    
\end{table}

\end{document}